\documentclass[runningheads]{llncs}

\usepackage{eccv}

\usepackage{eccvabbrv}

\usepackage{graphicx}
\usepackage{booktabs}

\usepackage{colortbl}
\usepackage{marvosym}

\usepackage[accsupp]{axessibility}  

\usepackage{hyperref}

\usepackage{orcidlink}

\begin{document}

\title{O$ _{2}$V-Mapping: Online Open-Vocabulary Mapping with Neural Implicit Representation} 

\titlerunning{O$ _{2}$V-Mapping}

\author{Muer Tie\inst{1,2} \and
Julong Wei\inst{1} \and
Ke Wu\inst{1} \and
Zhengjun Wang\inst{1} \and
Shanshuai Yuan\inst{1} \and
Kaizhao Zhang\inst{3}
Jie Jia\inst{1} \and
Jieru Zhao\inst{4} \and
Zhongxue Gan\textsuperscript{1\Letter} \and
Wenchao Ding\textsuperscript{1\Letter}
}

\authorrunning{M.Tie et al.}

\institute{Academy for Engineering \& Technology, Fudan University, China \\
\email{22210860058@m.fudan.edu.cn, \{ganzhongxue,dingwenchao\}@fudan.edu.cn}  \and
State Key Laboratory of Intelligent Vehicle Safety Technology\and
School of Future Technology, Harbin Institute of Technology, China
\and
Department of Computer Science and Engineering, Shanghai Jiao Tong University, China \\
}

\maketitle

\begin{abstract}
  %
  %
  %
  %
  %
  %
  Online construction of open-ended language scenes is crucial for robotic applications, where open-vocabulary interactive scene understanding is required. Recently, neural implicit representation has provided a promising direction for online interactive mapping. However, implementing open-vocabulary scene understanding capability into online neural implicit mapping still faces three challenges: lack of local scene updating ability, blurry spatial hierarchical semantic segmentation and difficulty in maintaining multi-view consistency. To this end, we proposed O2V-mapping, which utilizes voxel-based language and geometric features to create an open-vocabulary field, thus allowing for local updates during online training process. Additionally, we leverage a foundational model for image segmentation to extract language features on object-level entities, achieving clear segmentation boundaries and hierarchical semantic features. For the purpose of preserving consistency in 3D object properties across different viewpoints, we propose a spatial adaptive voxel adjustment mechanism and a multi-view weight selection method. Extensive experiments on open-vocabulary object localization and semantic segmentation demonstrate that O2V-mapping achieves online construction of language scenes while enhancing accuracy, outperforming the previous SOTA method.We have now open-sourced our code at \href{https://github.com/Fudan-MAGIC-Lab/O2Vmapping.git}{https://github.com/Fudan-MAGIC-Lab/O2Vmapping.git}.

  \keywords{NeRF \and Open-Vocabulary \and Language Scene Construction}
\end{abstract}

\section{Introduction}
\label{sec:intro}

\begin{figure}[htbp]
  \centering
  \includegraphics[width=\textwidth]{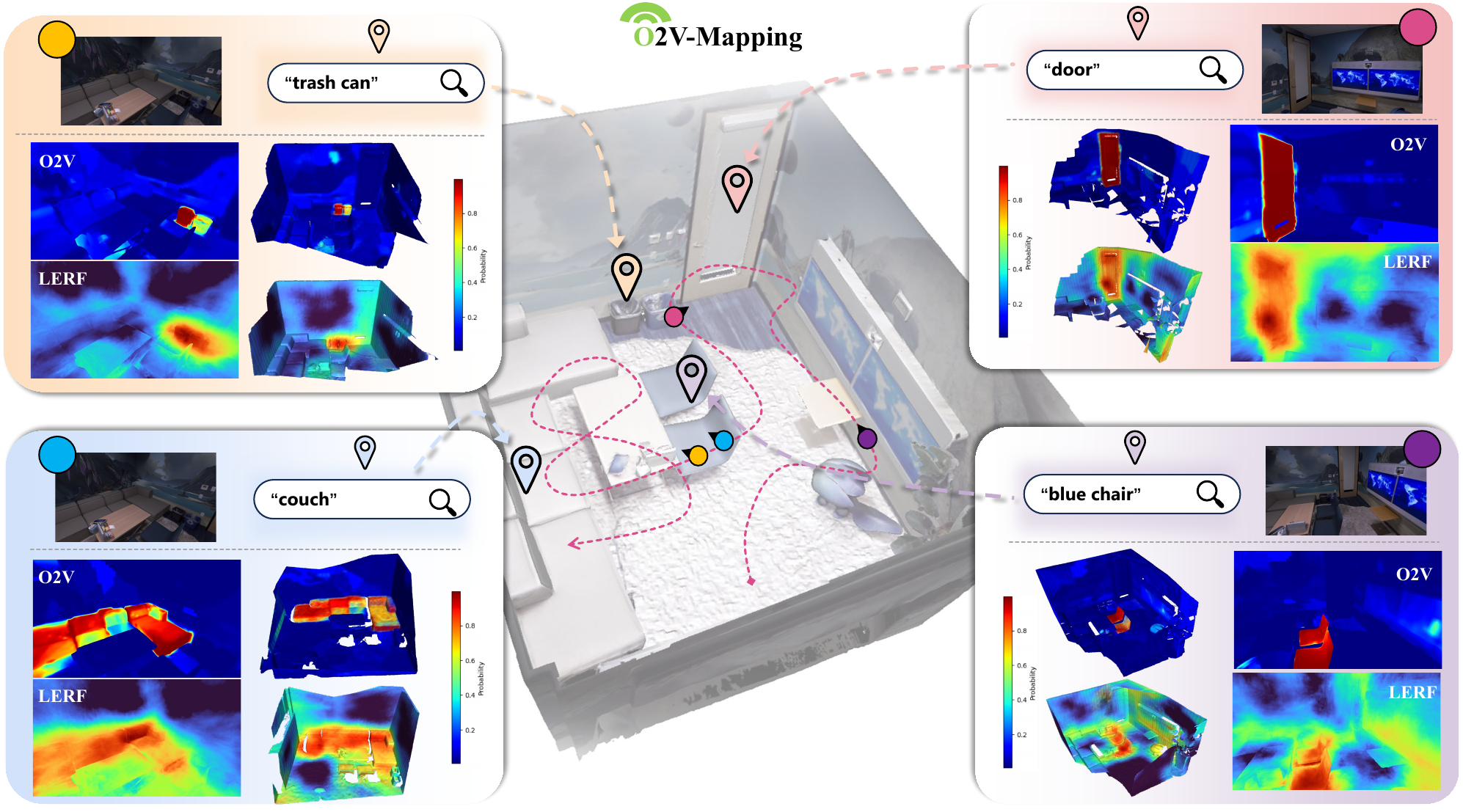}
  \caption{\textbf{Online Open-vocabulary Mapping.} O2V-mapping allows for online open-set text queries while constructing dense open-set semantic scenes, enabling the spatial localization of queried objects. Compared to LERF, it exhibits clearer object boundaries and more concentrated probability distributions of relevance.}
  \label{fig:example}
\end{figure}

Open-vocabulary scene understanding shows great potential in robotic applications such as navigation, object recognition, and human-robot interaction. Meanwhile, online mapping is indispensable for robots. Therefore, equipping the mapping framework with capabilities of open-vocabulary understanding and online reconstruction is highly meaningful for practical applications. 

NeRF-based implicit representation has been widely applied in mapping in recent years due to its high-fidelity reconstruction quality. However, solely utilizing Vanilla Neural Implicit Representation lacks semantic information. Existing semantic mapping methods primarily rely on manual annotation to obtain semantic information. However, this approach entails extensive workload and the obtained semantic information is limited by the finite number of semantic label categories. Open-vocabulary semantics offer the advantages of zero-shot acquisition of semantic information and unrestricted semantic category count, which is crucial for adapting to complex scenes. Recent techniques\cite{lerf2023} attempt to embed representationally open-set semantic language features such as CLIP\cite{radford2021learning} and DINO\cite{liu2023grounding} into neural radiance fields. They achieve open-vocabulary Novel View Synthesis (NVS) based on neural implicit representation. 

Nonetheless, aforementioned methods merely utilize heavy MLPs to represent the scene, leading to lengthy training time and they can only operate off-line on pre-collected data. Moreover, they introduce additional decoder branches to output semantic labels which adds extra burden to the framework. To adapt to complex and rapidly changing scenes, an online open-vocabulary mapping framework is significantly needed.

In terms of open-set semantic scene understanding, the common problem lies in the semantic information varying depending on different scales. For example, the object can be interpreted as the door and the door handle as our focus changes. Additionally, semantic vagueness is widely observed, where clear semantic boundaries cannot be obtained at the edges of objects. The key reason is that most existing works focus on pixel-level semantics, lacking object-level semantic understanding. Furthermore, during the process of online mapping, observing the same object from different perspectives may result in the assignment of different semantics to that object. The issues mentioned above contribute to spatiotemporal ambiguity in the process of semantic mapping.

In this paper, we present O2V-Mapping, a novel method for online constructing open-set semantic scenes efficiently and accurately. Our framework achieves online open-vocabulary mapping by grounding language embeddings from text-image large models like CLIP into voxel-based neural implicit representation. Moreover, by reconstructing open-set semantic scenes at the object level and utilizing segmentation priors of foundation models, we eliminate the semantic vagueness, get distinct semantic boundaries and resolve the multi-view inconsistency. Finally, we propose a LLM-centric agent architecture, instantiating our open-set semantic scene approach as the interactive memory, enabling full-scene grounding through query and render mechanisms to implement grounded tree search and online memory refinement. The main contributions of this work are summarized as follows. 

\begin{itemize} 

    \item We propose an online open vocabulary mapping framework by introducing a novel voxel-based open-vocabulary field (O2V Field). It enables online reconstruction as well as rendering high-fidelity RGBD and semantic images.
    \item We propose a language feature fusion mechanism to address the common issue of semantic spatiotemporal ambiguity in 3D scene understanding.

    \item We propose a LLM-centric agent architecture and achieve grounding of arbitrary objects within global scene.

\end{itemize}

\section{Related Work}

\textbf{\textit{Open-vocabulary 3D scene understanding.}} Recent advances in the field of open-vocabulary 3D scene understanding has been marked by the confluence of 2D visual language models and 3D point cloud processing techniques. The advent of Contrastive Vision-Language Pre-training (CLIP)\cite{radford2021learning} has notably enhanced the ability to comprehend scenes in a 2D open-vocabulary context without the need for semantic labels \cite{li2022languagedriven,liang2023openvocabulary,luo2023segclip,ma2023understanding,9896913,wu2024clipself,xu2023adapter,yu2023convolutions,zhang2023simple,zhong2021regionclip,zhou2022extract}. Several researchers\cite{conceptfusion} have effectively aligned the textual and visual features derived from  CLIP  with 3D point cloud features, facilitating the expansion of  open-vocabulary scene understanding to 3D scenes. Tschernezki et al.~\cite{tschernezki2022neural} introduced the DINO model \cite{liu2023grounding} for extracting 2D features, which are then subsequently utilized  for semantic feature learning within a trainable semantic field. Furthermore, Segment Anything Model (SAM)~\cite{kirillov2023segany} has demonstrated remarkable zero-shot capabilities across a multitude of downstream  2D tasks. Extending this innovation, SAM3D~\cite{yang2023sam3d} integrates the 2D masks generated by SAM with point clouds through projection, thereby imbuing the 3D point cloud space with semantic features. 

Additionally, Neural Radiance Field (NeRF)~\cite{mildenhall2020nerf} framework utilizes a continuous multi-layer perception (MLP) for implicit representation of three-dimensional scenes, offering advantages in terms of reduced storage requirements and enhanced continuity in scene modeling relative to traditional point cloud methodologies. Recent approaches have focused on how to utilize textual descriptions in conjunction with CLIP modeling to achieve NeRF-based implicit Open-vocabulary 3D scene understanding. For instance, Semantic NeRF \cite{Zhi:etal:ICCV2021} jointly encodes semantics with appearance and geometry within NeRF, leveraging NeRF's multi-view consistency and smoothness to accurately predict complete 2D semantic labels in specified scenes with only a small amount of specific semantic annotation data. LERF \cite{lerf2023} first uses CLIP features to supervise the formation of semantic scenes, supporting open-vocabulary queries within 3D scenes. Liu et al.~\cite{liu2024weakly} also utilize CLIP and DINO features to train a NeRF model for 3D open vocabulary segmentation. In comparison to these methods, our approach for the first time uses implicit features to construct 3D dense open-set scene understanding in an online manner. 

\noindent\textbf{\textit{Online implicit dense mapping.}} 
The field of online semantic 3D reconstruction is currently in its early stages, with existing research predominantly concentrating on 3D reconstruction and semantic understanding as separate entities. Revolutionary frameworks such as NeRF \cite{mildenhall2020nerf} and its derivatives \cite{rahaman2019spectral, müller2019neural}have transformed the creation of photorealistic images from limited viewpoints through the application of differentiable rendering. Acknowledging the extended training times required by coordinate encoding-based methods, a variety of studies \cite{yu2021plenoxels,Karnewar_2022,sun2022direct} have introduced the use of parameter encoding. This approach not only enlarges the parameter space but also significantly reduces the time needed for training. To improve the storage efficiency of parameter encoding-based techniques, innovations like octrees \cite{takikawa2021neural}, tri-planes \cite{chan2022efficient}, and sparse voxel grids\cite{li2023voxsurf, liu2021neural} have been developed. While a major focus of these techniques has been on synthesizing new views, other approaches have aimed at reconstructing the surfaces of RGB images using implicit surface models and differentiable rendering techniques\cite{guo2022neural, wang2023neus}. 

Advancements such as NICE-SLAM \cite{Zhu2022CVPR} have enhanced the scalability of conventional SLAM methods by incorporating neural networks. However, the combination of real-time semantic understanding with 3D reconstruction remains underexplored. Similarly, while Semantic SLAM \cite{semantic_slam} approaches have started to integrate semantic information to enhance the precision and usefulness of reconstructed models, the fusion of dynamic, real-time semantic interpretation with high-quality 3D reconstruction has not been fully achieved. Our research constitutes a pioneering attempt to address this deficiency, marking a good advancement in the domain of online semantic 3D reconstruction using neural implicit representation. 

\section{Methods}

\begin{figure}[ht]
  \centering
  \includegraphics[width=\textwidth]{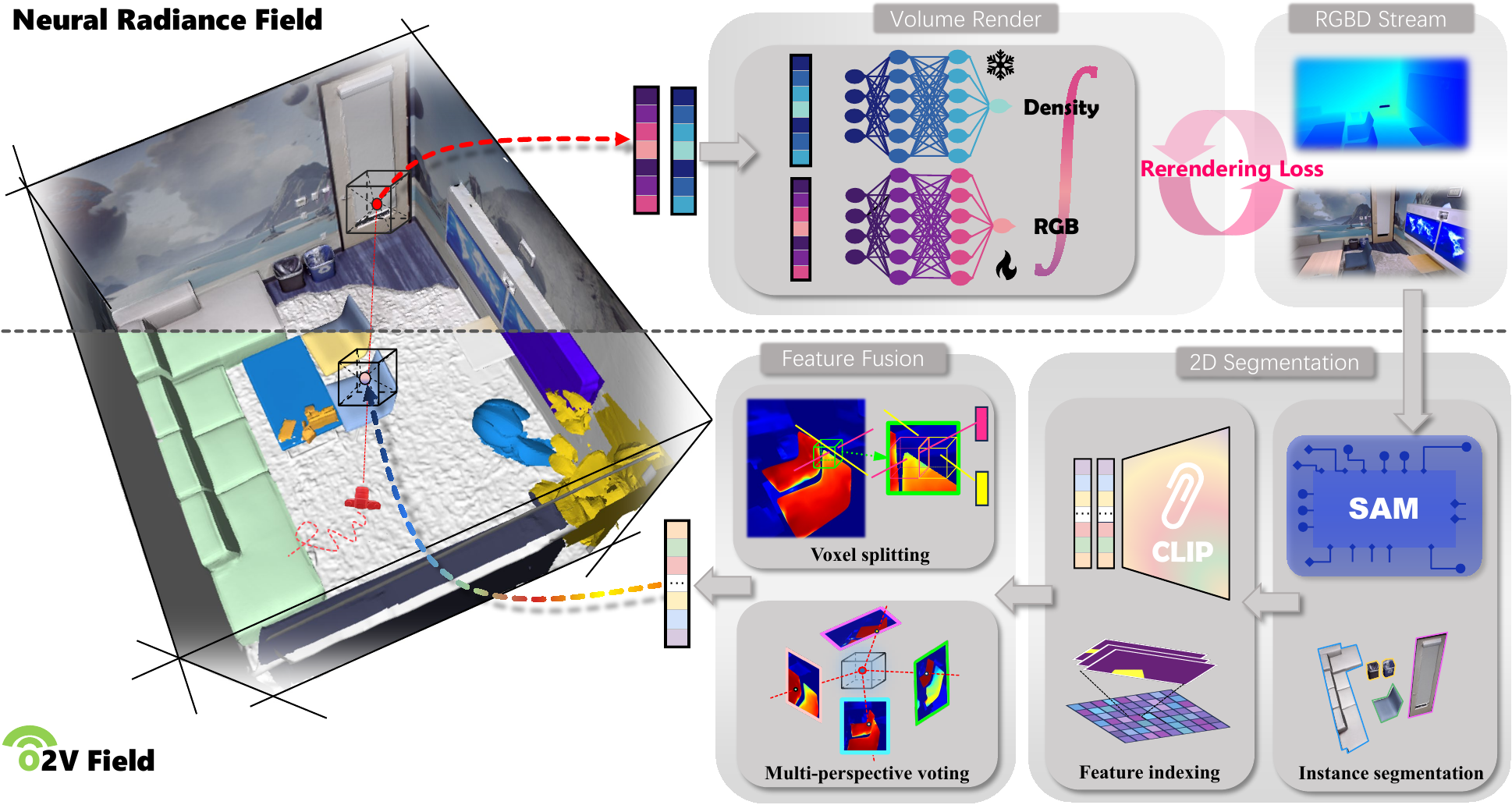}
  \caption{\textbf{The overview of our pipeline.} Top: Optimization of voxel-based neural radiance fields. Nearest trilinear interpolation is used to obtain color and geometric features for spatially sampled points. Then, leveraging NeRF's volume rendering, the scene is sampled and average-rendered to produce RGB and depth images. Bottom: Optimization of our O2V filed. We employ SAM to segment input RGB images and obtain instances. We further obtain language features through CLIP encoding. Feature indexing is performed to prepare for feature fusion. Finally, voxel splitting and multi-perspective voting are adopted to obtain fine-grained 3D open-vocabulary results.}
  \label{pipline}
\end{figure}

Our pipeline is illustrated in \cref{pipline}. An RGBD stream along with pose data are fed to our framework. We propose an object-level grounding method to extract CLIP embeddings effectively and incrementally integrate them into our O2V Field (\cref{text_Online_Construction_of_Open-Vocabulary_Field}). To alleviate the issues of hierarchical semantics and semantic ambiguity during online open-vocabulary mapping process, we propose a novel feature fusion mechanism to adjust and optimize our O2V field adaptively (\cref{text_Language_feature_fusion}).  Moreover, our O2V field can serve for a LLM agent to interact with the environment (\cref{text_Querying_O2V_field}). Our O2V field not only enhances the agent's ability to understand complex environments but also demonstrates promise in improving the adaptability and efficiency of LLM agents.

\subsection{Online Construction of Open-Vocabulary Field}\label{text_Online_Construction_of_Open-Vocabulary_Field}

\subsubsection{Voxel Implicit Representation with Multiple Categories.}

Current methods about constructing dense open-set semantic scenes such as LERF~\cite{lerf2023} mainly rely on resource-intensive neural networks to embed CLIP features. This restricts their ability to be updated online. To build an online open-vocabulary mapping pipeline, we employ hierarchical scene representations of various scales and types to store geometric and semantic information about scene objects.
Specifically, the scene is represented by three types of voxel grids: RGB, geometric, and semantic. 
These three distinct features are associated to each cell of the feature grid and updated in on online manner.
We utilize Multilayer Perceptrons (MLPs) to decode occupancy, color and semantic information from these features and perform volume rendering. 
Specifically, for a spatial point $p$,  we extract two fused latent vectors: the depth feature ${\phi}_d(p)$ and the color feature ${\phi}_c(p)$.
We use trilinear interpolation between the eight neighboring voxels to extract latent vectors. These vectors are concatenated with the position-encoded point $p$ and fed into two separate MLPs, $f^{d}$ and $f^{c}$, to decode occupancy probability and color values, respectively.

\begin{equation}
    o_{p} = f^d_{\theta}((p, \phi^d(p))), c_{p} = f^c_{\omega}(p, \phi^c(p)).
\end{equation}

The network structures of $f_{d}$ and $f_{c}$ follow the decoder design of ConvOnet \cite{tang2021saconvonet}.
$\theta$ and $\omega$ represent the trainable parameters of the two networks. Similar to \cite{10.1145/2508363.2508374}, we model the ray termination probability at point $p_{i}$ as 
    \begin{equation}
        w_{i} = o_{p_i} \prod_{j=1}^{i-1}(1-o_{p_j}).
    \end{equation}

The depth and color information rendered from ray tracing are predicted:
    \begin{equation}
                    \hat{D} = \sum_{i=1}^N(w_{i}d_{i}) , 
                    \hat{I} = \sum_{i=1}^N(w_{i}c_{i}).
    \end{equation}

For each ray, we compute the L2 loss between the predicted results and the ground truth depth and RGB values for $M$ sampled pixels:
    \begin{equation}
        \begin{gathered}
                L_{d} = \frac{1}{M}\sum^M_{m=1}(D_{m} - \hat{D_{m}})^2, \\
                L_{c} = \frac{1}{M}\sum^M_{m=1}(I_{m} - \hat{I_{m}})^2.
        \end{gathered}
    \end{equation}

Finally, we optimize the trainable parameters $\{\theta, \omega, \phi^d, \phi^c\}$ as follows : 
    \begin{equation}
        \min \limits_{\theta, \omega, \phi^d, \phi^c}(L_{d} + \lambda_{c}L_{c}).
    \end{equation}

For decoding semantic information, we utilize differentiable volume rendering to predict open-set semantic information from a given viewpoint. For a particular cell in the feature grid, it may have been observed from multiple view points and associated with distinct language features. To better utilize  multi-view observations, we set up a language feature queue for each cell. Each input language feature for the queue is associated with the corresponding mask confidence. Then we obtain the fused language feature $\hat{\textbf{f}}$ for a particular viewpoint by  

\begin{equation}
    \hat{\textbf{f}} = \frac{1}{||Q||}\sum^Q_{q=1}\sum_{i=1}^N(w_{i}f_{i}^q),
\end{equation}

where $f_{i}^q$ represents language features for multiple observations, $Q$ represents the full set of mask confidence queue, and $w_{i}$ represents the weight of points sampled near the predicted depth. This feature maintaining process is online and allows for text queries during the construction of the open-set semantic scene. It also enables open-set semantic tasks from novel viewpoints as well.

\subsubsection{Object-Level Open-Vocabulary Mapping.}


The input for CLIP are 2D images. However, the semantic composition of objects in real-world scenes is 3D and complex. Objects may have diverse semantic representations at different scales. What is worse, in 3D scene representations, a single-level image may contain multiple objects, leading to noise and confusion in the extraction of open-set semantic features.
To address these issues, we employ segmentation-based models to process the input image data.
Specifically, we first perform full-image segmentation on the input RGB image to obtain mask indices.
Utilizing the multi-scale functionality of the SAM model, we record the top three scores' mask indices in three separate queues for each pixel. 
Then, based on the instance segmentation images obtained on the 2D image, we employ a CLIP decoder to perform object-level encoding.
Finally, we project the obtained encoding into open-set semantic voxels based on depth information and store them in the feature queue.

\subsection{Language Feature Fusion}\label{text_Language_feature_fusion}

\subsubsection{Voxel Splitting.}\label{section3.2}

Due to the limitations of voxel-based representations in terms of resolution, conflicts in language features are inevitable when dealing with object boundaries in scenes, as well as objects with resolutions lower than the voxel resolution.
Increasing voxel resolution is an effective method, but may result in large computation overhead.

To this end, we propose an adaptive voxel splitting method, which involves splitting only selected voxels to avoid wasting storage space.
Specifically, thanks to object-level open-set semantic segmentation \cref{text_Online_Construction_of_Open-Vocabulary_Field}, when constructing open-set semantic scenes, language features for the same object are highly similar.
During the online update process, when multiple pixels are mapped to the same voxel with their cosine similarity exceeds a certain threshold, it indicates the presence of an object boundary or insufficient voxel resolution. In such cases, the voxel is uniformly split into eight higher-resolution voxels, each representing $1/8$ of the original volume. The language features are then stored separately based on the coordinates of the projection points. This adapative splitting strategy is critical to achieve online updating of open-vocabulary voxel grid.

In experiments, we split a voxel grid with side lengths of $16cm$ into eight voxel grids with side lengths of $8cm$. It is worth noting that we apply the voxel splitting strategy only to images from the same viewpoint. For images from different viewpoints, we employ multi-view voting to ensure consistency, which is introduced in the next section.

\subsubsection{Multi-view Voting.}

During the online construction of an open-set semantic scene, the challenge lies in achieving global consistency since information is obtained gradually over time in the form of RGBD streams. In cases where only a small portion of a 3D object is visible in 2D images, the confidence of CLIP results tends to be low in object-level open-set semantic mapping.
To address this issue, we employ a multi-view voting strategy to aggregate observations and increase the confidence of open-set semantic features for objects in scenarios where only a fraction of the object is visible across multiple viewpoints.
Specifically, we construct a feature queue to continuously observe a spatial object, adjusting the confidence levels of various features as the observation progresses. We use the cumulative observation count as weights to fuse the features, resulting in the following fused feature representation: 

    \begin{equation}
        \textbf{F}_{t}(\textbf{x})  = \frac{\sum(k_{t}(\textbf{x})\textbf{f}(\textbf{x}))}{{K}_{t}(\textbf{x})},
    \end{equation}

    \begin{equation}
        {K}_{t}(\textbf{x}) = \sum(k_{t}(\textbf{x})).
    \end{equation}
    
The rules for multi-view voting are as follows:

    \begin{equation}
        \textbf{F}^q_{t+1}(\textbf{x}) = \frac{K_{t}(\textbf{x})\textbf{F}^q_{t}(\textbf{x}) + k_{t+1}(\textbf{x})\textbf{f}^q_{t+1}(\textbf{x})}{K_{t}(\textbf{x}) + k_{t+1}(\textbf{x})},
    \end{equation}

    \begin{equation}
        K_{t+1}(\textbf{x}) = K_{t}(\textbf{x}) + k_{t+1}(\textbf{x}),
    \end{equation}
where $\textbf{F}^q_{t}(\textbf{x})$ and $K_{t}(\textbf{x})$ represent the language features and weight functions respectively, accumulated in the $q$ queue through voting from multiple images up to time $t$.


The confidence of features is positively correlated with the frequency of observations, meaning that the semantic understanding of an object relies on the weighted average of its various semantics across multiple viewpoints. This ensures that semantic information of objects remains stable throughout the online construction of an open-set semantic scene, thereby guaranteeing multi-view consistency.

\subsection{Querying O2V Field}\label{text_Querying_O2V_field}

\subsubsection{Establishment of Retrieval Map.}

According to \cref{text_Language_feature_fusion}, we can simultaneously construct an open-vocabulary neural implicit representation while conducting online text retrieval of 3D objects within the scene.
However, sampling all the voxels in the scene for objects that have already been sufficiently observed is extremely computationally expensive.
It is noted in the same section that the actual number of unique object features in a scene is significantly less than the number of voxels.
With high cosine similarity among features in voxels of the same object, we can establish a rudimentary open semantic map for the scene based on the types of object entities.
Specifically, we utilize the object masks obtained from the 2D images in \cref{text_Language_feature_fusion} to establish an index for the language features of each segmentation instance.
During the online construction of the open-vocabulary field, the number of indices keeps increasing, with most of them representing previously observed objects. Hence, we need to filter and merge them. We propose a criterion for determining whether merging is necessary:
\begin{equation}
    score = \frac{S(f^{t-1}_{i}, f^{t}_{j})}{distance_{(i, j)}},
\end{equation}
where $i$ and $j$ represent indices of different instances in the index, $S$ denotes cosine similarity, $f^t_{j}$ is the language feature of object $j$ at time $t$, and $distance$ represents the Euclidean distance between the geometric centers of the objects.
When $score > \alpha$ it is considered that they belong to the same object in space, and we update their language features and geometric centers accordingly. When querying the spatial position of an object, it is only necessary to perform CLIP on the indexed language features.

\subsubsection{Interacting with Large Language Model (LLM).}

\begin{figure}[!h]
  \centering
  \includegraphics[width=\textwidth]{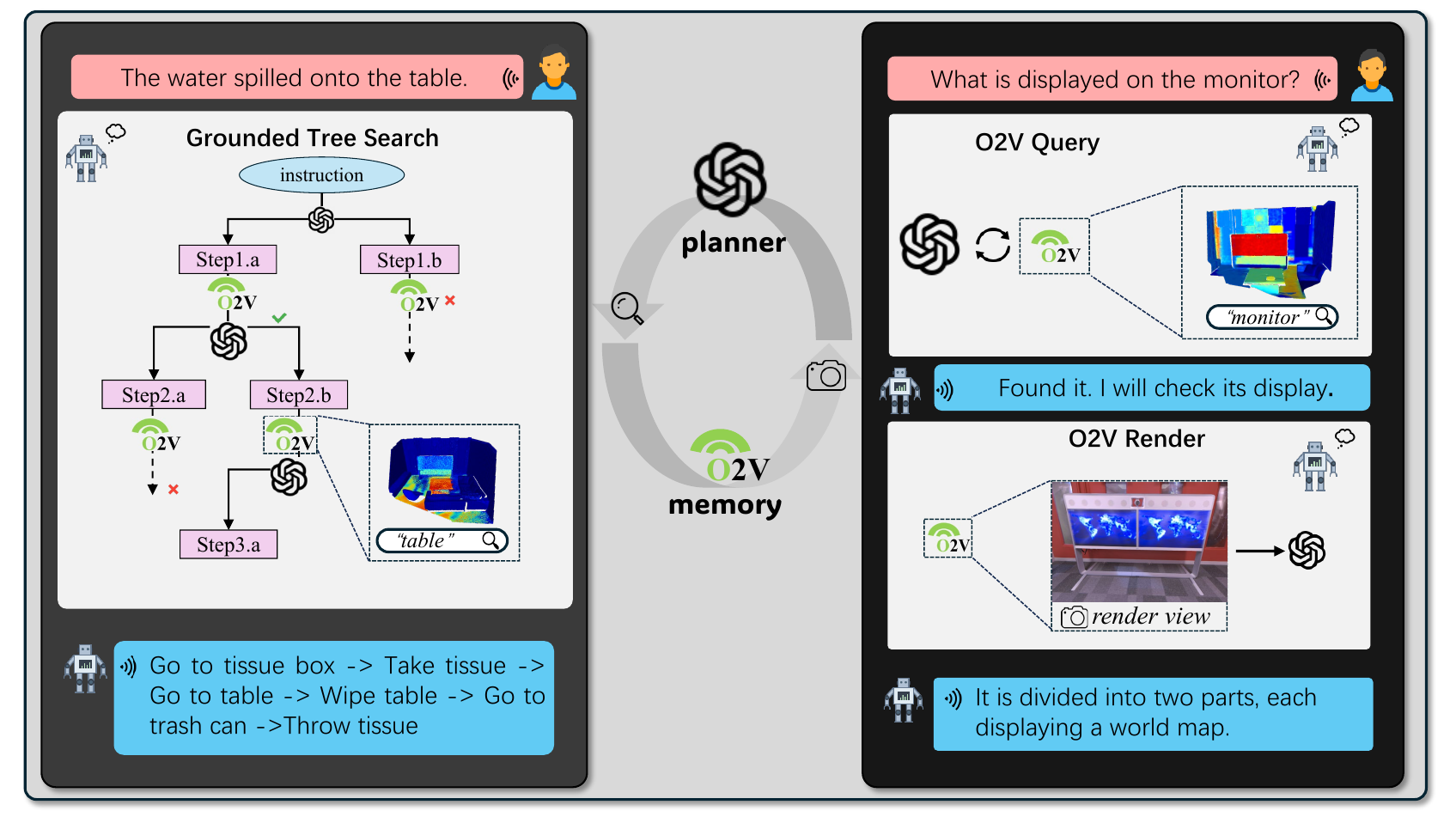}
  \caption{Left: The agent uses a search tree to roll out solutions, and optimizes the solution under the global scene based on feedback from the O2V map. For instance, at the "go to table" node, it queries the key object "table", which returns a high relevance, symbolizing the feasibility of this action. Right: For semantic content carried by the "display", it cannot be directly obtained through a query. Therefore, the agent first queries the carrier of semantic information and renders "diaplay" images. Then, the VLM can understand the content displayed is "world maps".}
  \label{fig:llm}
\end{figure}

The emergent inferential capabilities of LLM are impressive, while the lack of physical world understanding hindered its brilliance in robotic planning. Consequently, we explore the O2V map to provide global affordance for LLM-centric agent, aiming to enhance its grounding capability. As \cref{fig:llm} shown, this is achieved by instantiating O2V map as an interactive memory component for agent smoothly through detailed prompt engineering. It will open up query and render implementations to support grounded tree search and online memory refinement, respectively. For each node in grounded tree search, the agent will assess and optimize the planning affordance in the entire scene by querying the key objects involved. For cognitive information that cannot be directly obtained, the O2V map will render image of the information carrier's view. Then, the agent use VLM captioning capabilities to refine scene cognition online.

\section{Results}

In \cref{section4.1}, we qualitatively compared our relevance views with those of LERF~\cite{lerf2023} and analyzed the boundary segmentation of objects with different shapes and volumes.
In specific scenarios, LERF may exhibit a confused probability distribution of relevance during text queries, while our method ensures correct relevance rendering results. Our experiments were conducted on a single RTX 3090. The FPS of LERF was measured at 0.155. In contrast, our method achieved an FPS of 0.667, realizing a speed improvement over four times.
The analysis of these results is provided in \cref{section4.2}.
We conducted quantitative comparisons with other open-vocabulary methods such as OVSeg and LERF across numerous scenes. The results demonstrate that our method significantly outperforms the other two algorithms. Detailed analysis is provided in \cref{section4.3}. Furthermore, in our ablation studies~\cref{section4.4}, we detailed the critical importance of multi-view voting in preserving consistency and augmenting the stability of our framework. It is worth noting that, across all conducted experiments, only our approach succeeded in attaining online performance.

\label{sec:manuscript}

\subsection{Qualitative Results}\label{section4.1}

\begin{figure}[htbp]
  \centering
  \includegraphics[width=\textwidth]{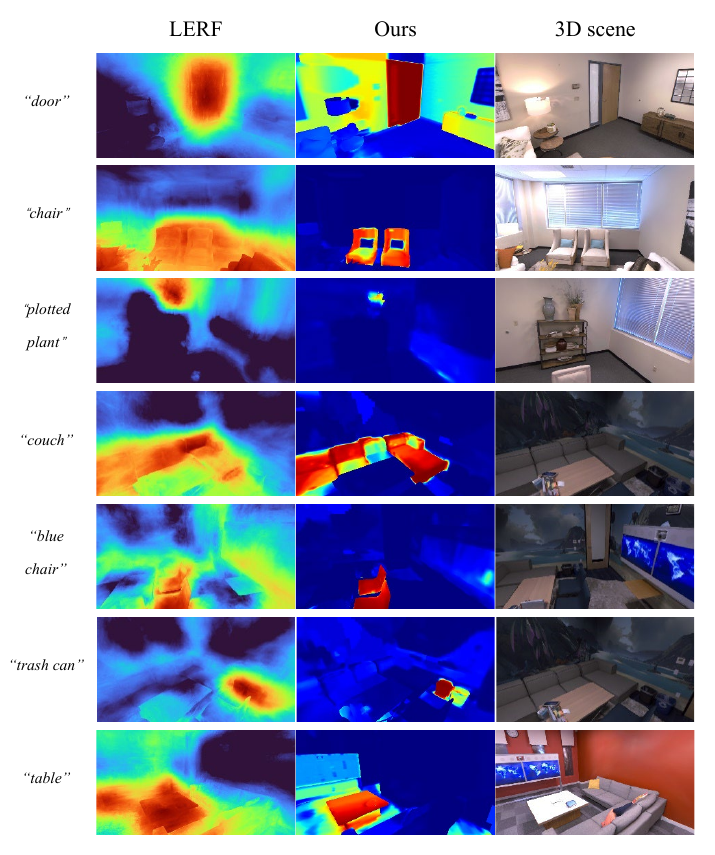}
  \caption{Our method is compared with LERF on 7 text query objects in 3 scenes. The relevance probabilities rendered by our method are concentrated around the queried objects, demonstrating clear boundary quality on objects of different shapes and sizes. Refer to \cref{section4.1} for a discussion and detailed information on relevance visualization.}
  \label{fig:comp}
\end{figure}

We utilize a normalization of maximum correlation similar to \cite{lerf2023} to visualize the results of text queries.
In \cref{fig:comp}, we selected 7 representative objects from 3 scenes from Replica\cite{replica19arxiv} to compare the natural language processing capabilities and the maximum correlation of queried objects with LERF.

\begin{figure}[h]
  \centering
  \includegraphics[width=\textwidth]{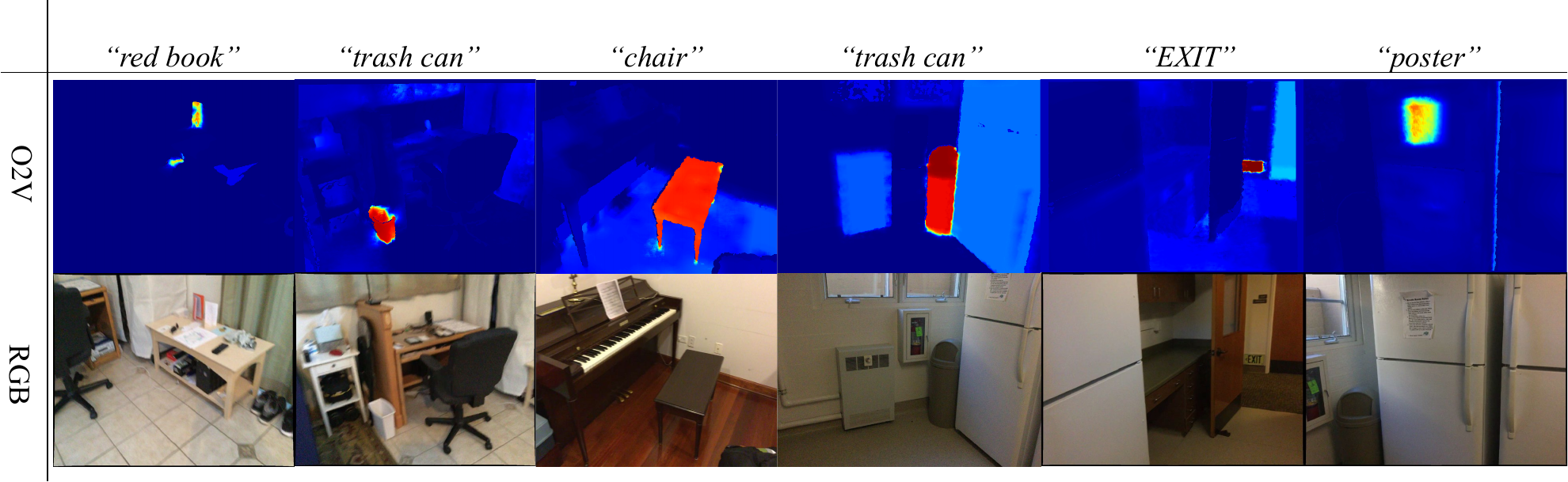}
  \caption{Our method was tested on three scenes and 18 object types from the ScanNet dataset, and it still performed well in real scenes with depth noise.}
  \label{fig:scannet}
\end{figure}

O2V-mapping not only allows for the online construction of open-set vocabulary scenes but also significantly improves object-level scene understanding, semantic segmentation accuracy, and precision compared to LERF.
For larger objects like "couch" that only appear in small portions in each image, O2V-mapping can still leverage the spatial consistency of implicit scene representations to understand the object as a whole, and demonstrate complete and clear object boundaries.
For small yet complex objects like "potted plant", O2V-mapping can also describe precise object boundaries.
Additionally, multiple objects of the same type, such as "chair" and "trash can" in the scene can also be assigned correct open-set semantic information by O2V-mapping. 
Even with slight differences in material, color, and shape among objects of the same type, they can still be clearly distinguished from surrounding objects of different types.
For objects like "table" and "door" with uniform textures and flat surfaces in the scene, O2V-mapping exhibits excellent performance. We also tested our method on the real-world ScanNet dataset with depth noise, and the results in \cref{fig:scannet} demonstrate that our method has good robustness.

\subsection{Query Robustness}\label{section4.2}

\begin{figure}[ht]
  \centering
  \includegraphics[width=\textwidth]{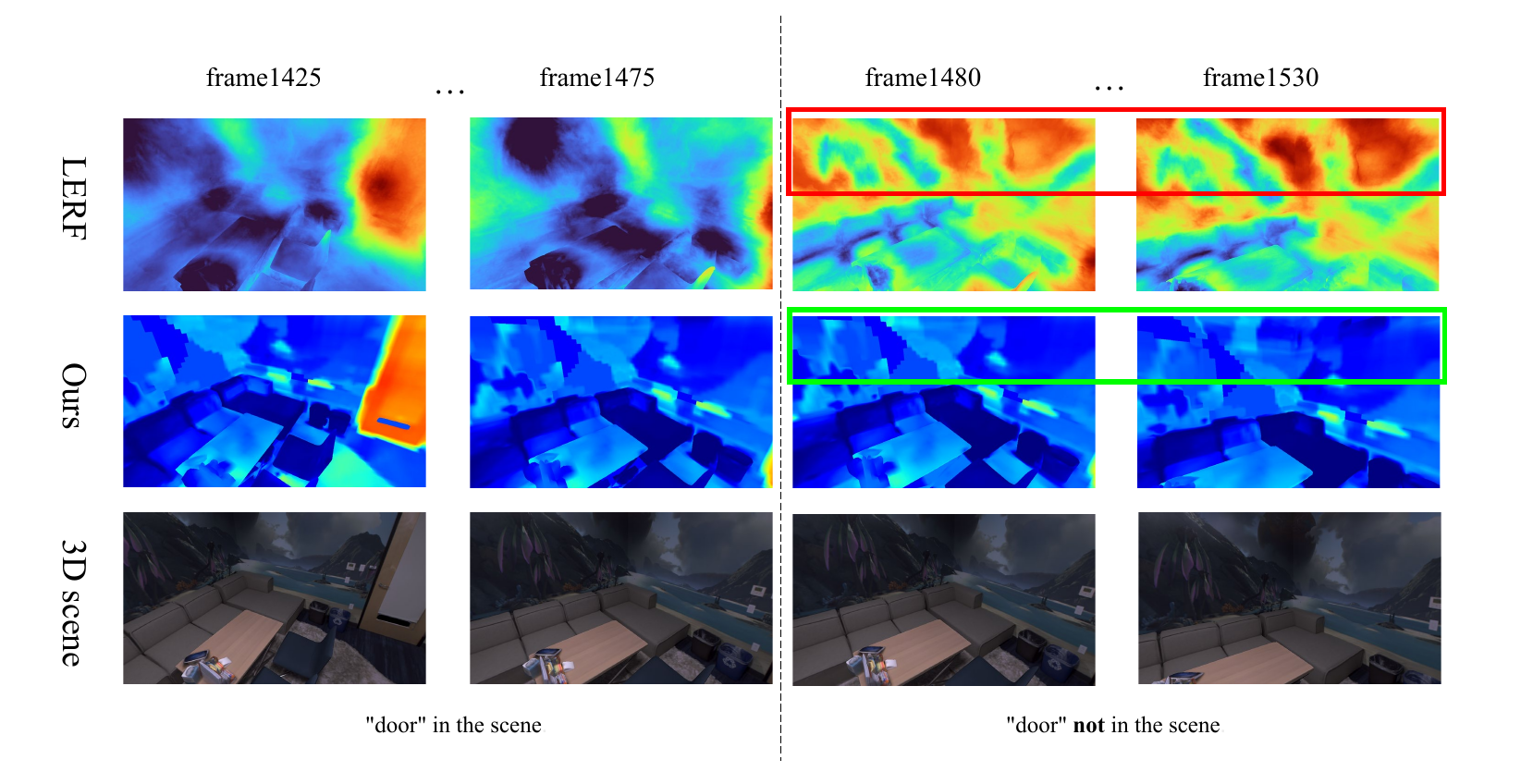}
  \caption{For a series of consecutive viewpoints in a scene, querying "door", LERF can roughly locate the object and render an approximate probability distribution of relevance when the queried object is present within the current viewpoint. However, when the queried object is not present within the current viewpoint, LERF exhibits a confused probability distribution of relevance, consistent with the sensitivity of LERF to query keywords mentioned in the referenced literature. Our method not only segments object boundaries well when only partial objects are observed but also maintains correct probability distributions of relevance even when the queried object ("door") is not within the field of view.}
  \label{fig:nodoor}
\end{figure}

LERF is highly sensitive to query keywords, making it prone to false positives when the query keywords are ambiguous or exhibit a "bag-of-words" behavior and struggles to capture spatial relationships between objects.
A common scenario is that when the queried object is not within the current viewpoint, LERF tends to exhibit a confused distribution of relevance scores (\ref{fig:nodoor}).
O2V-mapping exhibits robustness to query keywords. As shown in \ref{fig:nodoor}, in scenes where "door" is not present within the current viewpoint, there is no confused distribution of relevance scores.

\subsection{Open-set Segmentation and Localization}\label{section4.3}

\begin{table}
\centering
\caption{\textbf{Quantitative IoU metrics.} We compared the O2V map with 3D method LERF and 2D method OVSeg. We set up 4 rooms and calculated the IOU metric for 400 frame queries of 3 types of objects in each room to quantitatively evaluate. The results show that our method achieved the superior precision.}
\label{table1}
\begin{tabular}{l|ccc|cccc}
                     & \multicolumn{3}{c|}{Office0}                                                                                                            & \multicolumn{3}{c}{Office3}                                                                                                             &   \\ 
\hline
textquery            & "trash can"                                 & "blue chair"                                & "couch"                                     & "monitor"                                   & "table"                                     & "pillow"                                    &   \\
OVSeg                & {\cellcolor[rgb]{1,0.8,0.6}}0.6127          & 0.0105                                      & 0.0341                                      &    0.0364                                         & 0.3037                                      & {\cellcolor[rgb]{1,0.8,0.6}}0.5801          &   \\
LERF                 & 0.2234                                      & {\cellcolor[rgb]{1,0.8,0.6}}0.3817          & {\cellcolor[rgb]{1,0.8,0.6}}0.6537          & {\cellcolor[rgb]{1,0.8,0.6}}0.5679          & {\cellcolor[rgb]{1,0.8,0.6}}0.3480          & 0.3011                                      &   \\
Ours                 & {\cellcolor[rgb]{1,0.6,0.6}}\textbf{0.6257} & {\cellcolor[rgb]{1,0.6,0.6}}\textbf{0.8459} & {\cellcolor[rgb]{1,0.6,0.6}}\textbf{0.7563} & {\cellcolor[rgb]{1,0.6,0.6}}\textbf{0.6471} & {\cellcolor[rgb]{1,0.6,0.6}}\textbf{0.5983} & {\cellcolor[rgb]{1,0.6,0.6}}\textbf{0.6393} &   \\
\multicolumn{1}{l}{} &                                             &                                             & \multicolumn{1}{c}{}                        &                                             &                                             &                                             &   \\
                     & \multicolumn{3}{c}{Room0}                                                                                                               & \multicolumn{3}{c}{Room2}                                                                                                               &   \\ 
\hline
textquery            & "door"                                      & "chair"                                     & "window"                                    & "door"                                      & "window"                                    & "potted plant"                              &   \\
OVSeg                & 0.0291                                      & {\cellcolor[rgb]{1,0.6,0.6}}\textbf{0.3618} & 0.0775                                      &        0.0016                                     &        0.0328                                     &    0.0215                                         &   \\
LERF                 & {\cellcolor[rgb]{1,0.8,0.6}}0.1228          & 0.1756                                      & {\cellcolor[rgb]{1,0.8,0.6}}0.1232          & {\cellcolor[rgb]{1,0.8,0.6}}0.3143                      & {\cellcolor[rgb]{1,0.8,0.6}}0.2182                                      & {\cellcolor[rgb]{1,0.8,0.6}}0.2873                                      &   \\
Ours                 & {\cellcolor[rgb]{1,0.6,0.6}}\textbf{0.3090} & {\cellcolor[rgb]{1,0.8,0.6}}0.3118          & {\cellcolor[rgb]{1,0.6,0.6}}\textbf{0.3351} & {\cellcolor[rgb]{1,0.6,0.6}}\textbf{0.4290} & {\cellcolor[rgb]{1,0.6,0.6}}\textbf{0.5431} & {\cellcolor[rgb]{1,0.6,0.6}}\textbf{0.3942} &  
\end{tabular}
\end{table}

We present qualitative open-set segmentation results in \cref{table1}.
We can observe that our proposed method consistently outperforms existing approaches across all scenes in the Replica dataset and initialization settings.
Specifically, our method significantly improves mIoU by a factor of 1.12 compared to LERF\cite{lerf2023}, respectively. These results confirm the robustness and accuracy of our approach.
Furthermore, in multi-scene and multi-text query tasks involving different types of text queries, our method effectively utilizes single-view and cross-view information for voxel-based neural implicit representation-based 3D open-vocabulary segmentation.
We also evaluated the localization of objects queried by text.
\cref{table1} demonstrates that our object-level semantic embedding representations for object segmentation significantly outperform LERF and OVSeg in terms of localization accuracy in 3D scenes.

\subsection{Ablations}\label{section4.4}

\begin{figure}[h!]
  \centering
  \includegraphics[width=\textwidth]{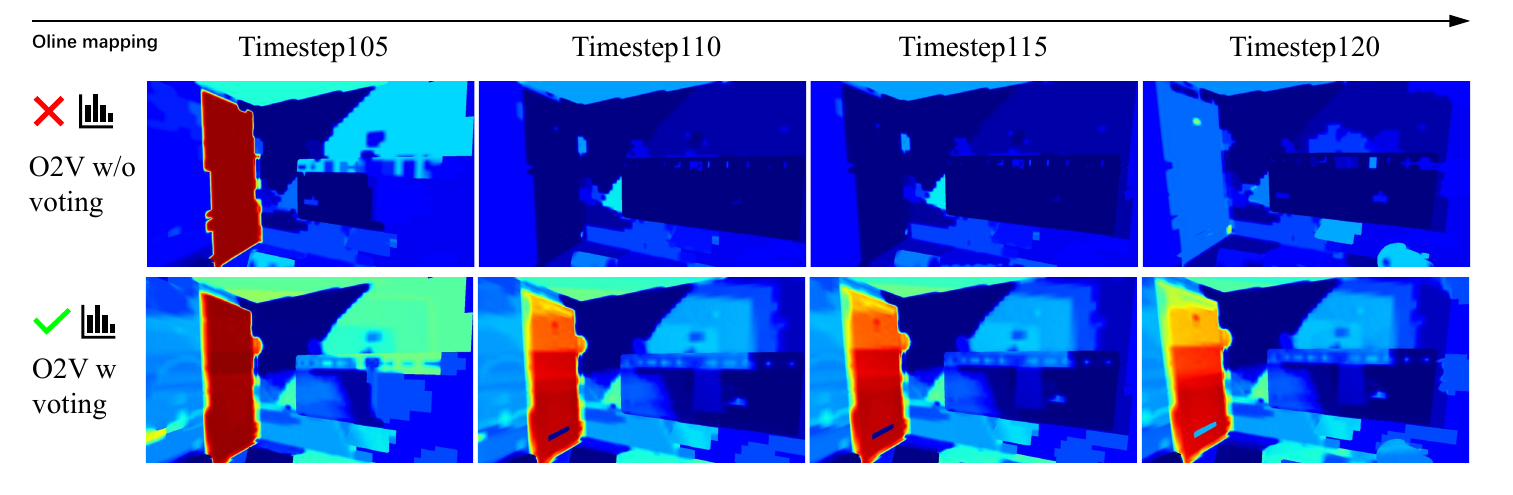}
  \caption{\textbf{Ablations:} We perform an ablation study by removing multi-view voting (\cref{section3.2}), and demonstrate here the errors in text queries caused by the degradation of the open-set semantic field during the online mapping process.}
  \label{fig:ablative}
\end{figure}

\textbf{Without multi-view voting:} Since 3D objects in space are not always entirely visible on the camera projection plane, the results segmented by SAM may not be accurately encoded by CLIP. This could result in some correct object parts or even all voxels being assigned incorrect semantic language features in frames with low confidence, leading to errors in text queries during the online process.
As shown in \cref{fig:ablative}, we demonstrate two illustrative examples where online text queries without multi-view voting may result in abrupt changes in the probability of the queried object, leading to query failures.
Multi-view voting, on the other hand, enhances the robustness of language features, preventing partial incorrect observations from affecting overall correctness.

\section{Conclusion}
To the best of our knowledge, we are the first framework to enable online construction of dense open-vocabulary fields using voxel-based methods and allowing online text queries.
Furthermore, extensive experiments demonstrate that our method for constructing open-set language fields significantly improves the clarity of object boundaries and exhibits highly robust text querying capabilities in 3D space, achieving state-of-the-art performance.
However, the limitations of the segmentation base model and the CLIP model restrict our method's text querying capabilities in certain specific scenarios.
Additionally, the interaction with some language models still appears somewhat rudimentary.
Therefore, there is still potential for further exploration in understanding arbitrary open-set semantic scenes and deeper language interactions.


%
%
\bibliographystyle{splncs04}

\section*{Acknowledgements}
This work is supported by Shanghai Pujiang Program (23PJ1400900) and State Key Laboratory of Intelligent Vehicle Safety Technology Open Fund Project (IVSTSKL-202317). Additionally, I extend my appreciation to the Shanghai Engineering Research Center of AI Robotics, Fudan University, China, and the Engineering Research Center of AI Robotics, Ministry of Education, China, for their invaluable support in the successful completion of this project.



\end{document}